\documentclass[a4paper]{article}
\usepackage{xspace}
\usepackage{todonotes}
\usepackage{devanagari}
\usepackage{INTERSPEECH2020}

\newcommand{\csreal}{{\textbf{CS-Real}}\xspace}
\newcommand{\hififty}{{\textbf{Hi-50}}\xspace}
\newcommand{\enfifty}{{\textbf{En-50}}\xspace}
\newcommand{\devset}{{\textbf{Dev}}\xspace}
\newcommand{\testset}{{\textbf{Test}}\xspace}

\title{Improving Low Resource Code-switched ASR \\ using Augmented Code-switched TTS}
\name{Yash Sharma$^1$, Basil Abraham$^2$, Karan Taneja$^1$, Preethi Jyothi$^1$} 
\address{
  $^1$Indian Institute of Technology Bombay, Mumbai, India\\
  $^2$Microsoft India Development Center, Hyderabad, India}
\email{\{yashsharma,karantaneja,pjyothi\}@iitb.ac.in, baabraha@microsoft.com}

\begin{document}

\maketitle
\begin{abstract}
Building Automatic Speech Recognition (ASR) systems for code-switched speech has recently gained renewed attention due to the widespread use of speech technologies in multilingual communities worldwide. End-to-end ASR systems are a natural modeling choice due to their ease of use and superior performance in monolingual settings. However, it is well-known that end-to-end systems require large amounts of labeled speech. In this work, we investigate improving code-switched ASR in low resource settings via data augmentation using code-switched text-to-speech (TTS) synthesis. We propose two targeted techniques to effectively leverage TTS speech samples: 1) Mixup, an existing technique to create new training samples via linear interpolation of existing samples, applied to TTS and real speech samples, and 2) a new loss function, used in conjunction with TTS samples, to encourage code-switched predictions. We report significant improvements in ASR performance achieving absolute word error rate (WER) reductions of up to 5\%, and measurable improvement in code switching using our proposed techniques on a Hindi-English code-switched ASR task.  
\end{abstract}


\section{Introduction}

Code mixing or code switching (CS) is a widespread linguistic phenomenon among speakers in multilingual communities, where they switch between two or more languages often within the confines of a sentence. Code switching appears naturally in conversational speech. With the prevalence of social media, code switching also appears frequently in textual form. As a result, building computational models for code switched speech and text has gained a lot of recent attention.

As is typical for state-of-the-art ASR systems, CS speech ASR would also require large quantities of labeled speech during training. While one could reasonably hope to gain access to large monolingual corpora in the component languages, CS speech data is a relatively scarce resource. Thus, it is of great interest to investigate ways in which one can augment CS speech data in order to effectively train CS ASR systems.

In this work, we investigate the effective use of CS text-to-speech (TTS) samples to augment ASR systems for CS speech. With CS text as the starting point, large quantities of TTS samples can be generated. TTS systems can be trained with far fewer hours of clean labeled speech compared to ASR systems. However, there is a distributional mismatch between TTS samples and real speech samples. Using TTS samples as-is could lead to over-fitting on artefacts in the TTS data. The main question of interest in this work is how to best leverage a fixed quantity of TTS samples to improve ASR for CS speech. Towards this, our contributions can be summarized as follows:
\begin{itemize}
    \item We suggest mixup~\cite{zhang2017mixup} as an appropriate strategy to be applied to TTS samples and real speech samples in order to help reduce the distributional mismatch.
    \item We propose a new \emph{CS-bias} loss function that encourages the ASR system to code-switch and performs well in conjunction with TTS samples.
    \item We demonstrate the efficacy of our proposed techniques on a Hindi-English code-switched ASR task. We also measure error reductions at the switching points and observe clear improvements in performance.
\end{itemize}

\section{Related Work}


Prior work on speech recognition for CS speech has focused on aspects relevant to both the acoustic model and the language model. For the acoustic model, previous approaches have investigated efficient ways to do phone merging~\cite{seame}, tying mixture models of two languages within a GMM-HMM model~\cite{asymmetric-cs-asr} and using language-dependent inference~\cite{univphoneme-mixed-asr}. There has also been a lot of prior work on improving LMs for CS text that includes the use of parallel text in the CS language pair~\cite{hkust-pointer-paper}, the use of generative models to synthesize CS text~\cite{FAME-paper,adel2013recurrent,csalt-dlm-speech-paper,ect_paper}. In more recent work, end-to-end models have also been explored for code-switched ASR with improvements observed 
using language identification \cite{e2e-manglish-icassp19,cm-asr-ctc-icassp19,multilingual-and-cm-google-paper} in multitask (MTL) settings,
jointly using CTC and attention based loss \cite{kim2016joint,srivastava2019endtoend} within an MTL model,
multi-graph decoding and LM rescoring \cite{Yue2019EndtoEndCA} for low-resource CS ASR, 
and using various units like wordpieces \cite{e2e-manglish-icassp19} and byte-pair encoding (BPE) units~\cite{google-bytes-paper} to model text.
Apart from techniques targeted at CS ASR, several other data augmentation techniques have been shnown to boost the performance of ASR systems such as speed perturbation \cite{Ko2015AudioAF}, randomized spectrogram masking \cite{SpecAugment}, vocal tract length perturbation and semi-supervised learning \cite{Ragni2014DataAF}.

Due to the lack of sufficient CS labelled corpora for training, \cite{merl-concat-paper,karan_paper} train ASR systems using synthetic datasets created by concatenating monolingual segments from the constituent languages of the CS speech. 
In contrast, we explore the use of samples generated from a TTS systems for training our ASR system.
\cite{google-aug-tts,wang2020improving} use TTS samples as an augmentation technique to improve ASR performance on the LibriSpeech corpus.   
For monolingual speech recognition in the low-resource scenario, \cite{tts-aug-monolingual} uses TTS samples to introduce speaker diversity.
We propose techniques that utilize TTS speech for code-switched ASR with access to only 12 hours of real CS speech. 
\cite{nara-chain-paper} uses TTS and ASR set up as a chain to train a CS ASR system from non-parallel speech and text corpora.
Though synthetic CS data can be generated using synthesis engines~\cite{jap-eng-tts-data,kano2018structuredbased}, TTS samples significantly differ from real speech in their acoustic characteristics and hence need to be suitably handled in order to be more effective as sources of data augmentation.

\section{Our Proposed Techniques} \label{setup:main}

The use of TTS samples for data augmentation and improving ASR performance has been investigated in recent work~\cite{google-aug-tts,wang2020improving}. In order to insulate the ASR model from artefacts specific to TTS data, prior work has observed that freezing the encoder parameters \cite{DBLP:journals/corr/abs-1811-02050} and only updating the rest of the model using TTS samples is beneficial. We present this strategy as a baseline and propose two other techniques in this section that further improve performance. 


\subsection{Mixup Algorithm and Sampling} \label{setup:mixup}

Mixup~\cite{zhang2017mixup} was originally proposed as a simple regularization technique to train a classifier using convex combinations of pairs of training instances and their corresponding labels. Mixup has also been investigated within the paradigm of ASR~\cite{Medennikov2018AnIO,saon2019sequence,zhu2019mixup,liang2019levoice}. Given spectral features of two speech samples, $X_a$ and $X_b$, the \emph{mixed} training example $X_{\text{mix}}$ is given by: 
\begin{equation}
    X_{\text{mix}} =   \lambda_{\text{mix}} \cdot X_a + (1 - \lambda_{\text{mix}}) \cdot X_b
\end{equation}
Here, $\lambda_{\text{mix}} = \max(\lambda, 1-\lambda)$ where $\lambda \sim \text{Beta}(\alpha, \beta)$, and $\alpha$, $\beta$ are the standard shape parameters of the beta distribution. We apply this mixup strategy to TTS samples and real speech samples such that $X_a$ is a TTS sample and $X_b$ is a randomly-chosen real speech sample. For our experiments, we set $\alpha=0.4$ and $\beta=0.4$. They were fine-tuned on \devset set. The resulting shape of the beta distribution ensures that a $\lambda$ sampled from this distribution would either be very close to 0 or 1. Given this skew, we set the transcription of the mixed utterance $X_{\text{mix}}$ to be the same as that of the TTS sample $X_a$. 

With the above-mentioned mixup strategy in place, we intend for real speech samples to be mixed in with TTS samples to bring the latter a little closer to real speech in the input feature space. Given that the acoustic properties of TTS speech differ from natural speech in various dimensions, particularly in its intonational properties, we hope mixup helps mitigate these differences to a small extent and acts as a regularizer. This is indeed borne out by our experiments shown in Section~\ref{sec:mixup}.

%
%




\subsection{Rewarding Code-switching with a \emph{CS-bias} Loss} \label{setup:reward_in_loss}
%
%
Before describing our new loss function, we briefly outline the end-to-end ASR architecture employed in this work. Our base ASR model is a standard hybrid CTC-attention end-to-end architecture that uses an encoder-decoder paradigm with attention~\cite{watanabe2018espnet}. This model is trained with a multitask objective that uses a weighted combination of a CTC-based loss and an attention-based loss via a shared encoder. The combined loss function is referred to as $L_\text{MTL}$. For CS speech, we devise a new loss term to be added to $L_\text{MTL}$ that explicitly rewards code switching. 

Many common English words have been assimilated into the Hindi language and appear in the Devanagari script within the Hindi vocabulary. They also appear as English words in their  romanized forms. However, since Hindi is typically the dominant (matrix) language in Hindi-English CS speech, the model is hesitant to code-switch even when the acoustics have sufficient evidence of a potential switch point. Our proposed modification to the loss function rewards higher probabilities being assigned to English characters. This reward can be quantified as: $\mathbf{R} = \displaystyle \sum_{t = 1}^{N} \sum_{c \in \mathcal{S}_e} P_t(c)$
, where $P_t(c)$ is the probability at output timestep $t$ assigned to a character $c$ that is drawn from the set of all English characters denoted by $\mathcal{S}_e$. (The output sequence is of total length $N$.)
%
%
%
%

The modified \textit{CS-bias} loss function then becomes: 
\begin{equation}
    L_{\text{CS}} =  L_{\text{MTL}} - \lambda' \cdot (\mathbf{R}_{\text{CTC}} + \mathbf{R}_{\text{ATT}})
\end{equation}
where $\mathbf{R}_{\text{CTC}}$ and $\mathbf{R}_{\text{ATT}}$ are rewards specific to the CTC and attention outputs and $\lambda'$ is the scaling coefficient associated with the new loss term. The reward term is meant to serve as an explicit boost to English characters whenever speech switches to English.  



\section{Experimental Setup}

\paragraph*{Datasets.}
We use a proprietary dataset from Microsoft comprising 50 hours of Hindi monolingual speech (\hififty) and 50 hours of Indian accented English speech (\enfifty).%
%
We also make use of  approximately $12$ hours of real CS Hindi-English speech that amounts to 12K CS utterances (i.e. 12K different transcriptions). This dataset will henceforth be referred to as \csreal. The speech corpus consists of mostly conversational type utterances collected in a multi-condition scenario. The utterances are transcribed in both Roman and Devanagari scripts corresponding to English and Hindi language respectively ~\cite{karan_paper}. For the first set of experiments, until Section~\ref{setup:reward_in_loss}, we restrict ourselves to TTS samples generated using the text in \csreal and assume no access to real CS speech. In Section~\ref{results:skyline}, we show the effect of our techniques on a model that has been finetuned using speech in \csreal. 

Our models are evaluated on two evaluation datasets comprising real CS speech. The development set (\devset) consists of $2,000$ sentences with a total of $25,679$ words and the unseen test set (\testset) contains $2,000$ sentences with $29,408$ words overall. \devset is used to tune $\alpha$, $\beta$, and other hyperparameters that appear in our techniques.


\paragraph*{TTS Generation.}
To generate TTS samples, we train a Tacotron-2 model~\cite{tacotron2} that employs a multi-speaker, multilingual acoustic model and maps phone sequences to Mel spectrograms. Our model is trained using roughly 30,000 utterances from six different Indian languages -- English, Hindi, Tamil, Telugu, Gujarati and Marathi -- spoken by a single speaker in each language. The acoustic model is created using a common phone set which is the union of phone sets of all six languages. Phones that are common across different languages get merged, while phones that are specific to a language remain intact.We use a parallel Wavenet vocoder~\cite{Vocoder} that is trained on large set of Microsoft proprietary TTS data.


We train the acoustic model for 250K iterations with a batch size of 32. We also make use of high-quality speech data from 70 Hindi speakers speaking in different accents from a larger Hindi dataset from which \hififty was drawn~\cite{karan_paper}. We further adapt our acoustic model with 100-200 samples from specific speakers for an additional 10K iterations with the same training hyperparameters. This final adaptation step yields speaker-specific models for each speaker. Finally, we feed CS text from \csreal as input and generate speech samples from these speaker-adapted models that amount to roughly 100 hours of speech overall. 

\paragraph*{Implementation Details.} 
We use the ESPnet toolkit~\cite{watanabe2018espnet} for all our  experiments as it provides a flexible interface for many end-to-end (E2E) architectures.
As mentioned in Section~\ref{setup:reward_in_loss}, we train our baseline model using a multi-task (MTL) objective that combines a CTC loss and an attention loss via a shared encoder. All our models use graphemes as  the output vocabulary.%
\footnote{We also tried a byte-pair-encoding (BPE)~\cite{sennrich2015neural} based model, but this did not perform as well as the purely grapheme-based models.}
%


The encoder architecture consists of 4 layers of stacked bidirectional LSTMs, with 512 encoder units, 512-dimensional encoder projections and sub-sampling to reduce sequence lengths. 
We used a location-aware attention-based model with a 1-layer decoder for the CTC and attention modules each.
The maximum input length is 800 frames and the maximum output length is 150 tokens. $\lambda_{\text{MTL}}$ is the interpolation coefficient used in the combined loss $L_\text{MTL}$; this was set to $0.7$ in all our experiments, inspired from results in \cite{srivastava2019endtoend}
%
%
%

Our output vocabulary includes both English and Devanagari characters, along with a ``space" character, an unknown marker and an end-of-sentence marker. 
The model was trained using the Adadelta optimizer~\cite{zeiler2012adadelta} with a base learning rate of 1.0. We trained the network for a total of 20 epochs with a patience value set to 10.
All experiments used data shuffling as well as sortagrad~\cite{amodei2015deep} for all epochs, 
i.e. we first sorted the data by length, batched the samples, and then shuffled the order of batches. 
This turned out to perform better than doing no sortagrad and no shuffling. The effective batch size was set to 32 in all our experiments.
For some experiments where we fine-tune a pre-trained model using TTS samples, we freeze the encoder as described in \cite{DBLP:journals/corr/abs-1811-02050}. These experiments are labelled with \textbf{FE} (frozen encoder).

We used a word-based recurrent neural network language model (RNNLM) with a weight of 0.1. The RNNLM is a single-layer LSTM with 1000 units and an output vocabulary restricted to the 2000 most frequently occurring words in \textbf{CS-Real}. 
It is trained for 40 epochs before converging with a patience of 20. 
%
%
During decoding, we used beam decoding with a width of 20 and no insertion penalty was invoked. 
%


%



\section{Results}

Our baseline hybrid CTC-based ASR model is trained on the two monolingual \hififty and \enfifty datasets merged together.%
\footnote{By tuning on \devset, we observed that merging the Hindi and English monolingual data in unit ratio performed the best.}
The subsequent sections will be structured as follows. In Section~\ref{sec:mixup}, we show the benefits of mixup in conjunction with TTS samples. Section~\ref{sec:csbias} will show the effect of using the \emph{CS-bias} loss function when using TTS samples. Finally, in Section~\ref{results:skyline}, we finetune our baseline ASR system using real CS speech in \csreal and examine the effects of our proposed techniques on the resulting system.

\subsection{Mixup with TTS samples}
\label{sec:mixup}

Table~\ref{tab:tts_mixing} shows how mixup is beneficial when used in conjunction with TTS samples. As mentioned before, the ``Baseline" system does not make use of any CS speech. The ``no-mixup" systems refer to the use of TTS samples with the real speech samples without any mixup. The numbers alongside each system denote the number of hours of TTS speech used in each experiment. That is, ``no-mixup 50" refers to the use of 100 hours of real speech (from \hififty and \enfifty) and $50$ hours of TTS speech without any mixup in place. We observe that adding code-switched TTS samples to the monolingual speech, even without any mixup, is beneficial. Adding more TTS data improves performance further, however there is a trend of diminishing returns ($72.9 \rightarrow 71.4 \rightarrow 71.0$). The ``add-mixup" systems, in contrast to ``no-mixup",  uses mixup with each batch of TTS samples. Each batch of TTS speech samples is mixed with a different batch of real speech, and $\lambda_{mix}$ is sampled anew for each such batch-wise mixing. We observe clear improvements in performance with using ``add-mixup" in comparison to ``no-mixup". 
\begin{table}[t!]
    \caption{WERs (CERs) using TTS data, with and without mixup}
    \centering
    \begin{tabular}{l c c}
    \toprule
    \textbf{System}  & \devset & \testset \\
    \midrule
    Baseline & 70.7 (54.3) & 72.1 (54.5) \\
    \midrule
    no-mixup 20  & 71.5 (49.1) & 72.9 (49.2)\\
    no-mixup 50 & 69.7 (48.6) & 71.4 (48.5)\\
    no-mixup 100 & 70.0 (48.7) & 71.0 (48.2)\\
    \midrule
    add-mixup 20 & 70.3 (48.5)& 71.4 (48.2)\\
    add-mixup 50 & \textbf{68.8} (47.9) & 70.3 (47.9) \\
    add-mixup 100 & 69.0 (47.9) & \textbf{70.0} (\textbf{47.4})\\
    \bottomrule
    \end{tabular}
    \label{tab:tts_mixing}
\end{table}
\begin{table}[t!]
    \caption{WERs (CERs) using TTS data, with and without freezing the encoder}
    \centering
    \begin{tabular}{l c c}
    \toprule
    \textbf{System}  & \devset & \testset \\
    \midrule
    no-mixup 100 & 70.0 (48.7) & 71.0 (48.2)\\
    no-mixup 100 (\textbf{FE}) & 71.2 (46.7) & 73.0 (47.6) \\
    \midrule
    add-mixup 100 & \textbf{69.0} (47.9) & \textbf{70.0} (47.4)\\
    add-mixup 100 (\textbf{FE}) & 69.5 (\textbf{45.2}) & 71.3 (\textbf{46.3}) \\
    \bottomrule
    \end{tabular}
    \label{tab:tts_mixing_FE}
\end{table}

We examine the effect of freezing the encoder (\textbf{FE}) as suggested in~\cite{DBLP:journals/corr/abs-1811-02050} on both ``no-mixup" and ``add-mixup" in Table~\ref{tab:tts_mixing_FE}. We observe that freezing the encoder was not an effective strategy in our setting possibly due to the use of smaller amounts of TTS data (in comparison to the corpus used in prior work). However, the CERs do improve with freezing the encoder. This will need further investigation which we leave as future work.

When augmenting real speech samples with TTS samples in the ``add-mixup" setting, there are two ways in which this could be implemented. The mixup samples could be used to train the ASR system from scratch or one could start with the baseline ASR system and further finetune it using the mixup samples. Table~\ref{tab:finetune_or_scratch} shows results from both these training strategies.
\begin{table}[b!]
    \caption{WER(CER): \textit{Finetune} vs \textit{from scratch} experiments}
    \centering
    \begin{tabular}{l c c}
    \toprule
    \textbf{System}  & \textbf{DEV} & \textbf{TEST}  \\
    \midrule
    add-mixup 20 (scratch) & 70.3 (48.5)& 71.4 (48.2)\\
    add-mixup 50 (scratch) & \textbf{68.8} (47.9) & 70.3 (47.9) \\
    add-mixup 100 (scratch) & 69.0 (47.9) & \textbf{70.0} (47.4)\\
    \midrule
    add-mixup 20 (\textbf{FT}) & 69.8 (48.4)& 71.6 (48.3)\\
    add-mixup 50 (\textbf{FT}) & 69.7 (47.6)& 70.9 (47.7)\\
    add-mixup 100 (\textbf{FT}) & 68.9 (48.0)& 70.2 (47.8)\\
    \bottomrule
    \end{tabular}
    \label{tab:finetune_or_scratch}
\end{table}
Training from scratch appears to be a more effective strategy yielding lower error rates on both \devset and \testset compared to the systems using finetuning (denoted by \textbf{FT}).

\subsection{Influence of \emph{CS-bias} loss function}
\label{sec:csbias}
We observe that the modified \emph{CS-bias} loss function is effective in encouraging CS predictions and reduces the overall WER on our evaluation datasets, over and above the improvement gained by using mixup. Explicitly rewarding English predictions helps overcome the model's bias towards predicting back-transliterated Devaganari words instead of English words. In all our experiments involving \emph{CS-bias}, $\lambda'$ was set to $0.25$.
\begin{table}[t!]
    \caption{WERs (CERs) highlighting the effect of CS-bias. A-100 refers to the ``add-mixup 100" system.}
    \centering
    \begin{tabular}{l c c}
    \toprule
    \textbf{System}     & \devset & \testset \\
    \midrule
    Baseline & 70.7 (54.3) & 72.1 (54.5) \\ 
    A-100 & 69.0 (47.9) & 70.0 (47.4)\\ 
    A-100 (\textbf{FT})  & 68.9 (48.0)& 70.2 (47.8)\\
    \midrule
    A-100 + \emph{CS-bias} (scratch) &  \textbf{67.7} (47.4) & \textbf{69.0} (47.3)\\ 
    A-100 (\textbf{FT}) + \emph{CS-bias} & 68.6 (47.3)& 70.2 (47.3)\\
    \bottomrule
    \end{tabular}
    \label{tab:boosting}
\end{table}

Table~\ref{tab:boosting} shows how \emph{CS-bias} influences performance. Including this loss term during training of an ``add-mixup 100" system from scratch is most beneficial. We see a clear reduction in WERs both on the \devset and \testset datasets. We also tried including the loss term as part of the finetuning phase while training ``add-mixup 100 \textbf{FT}". However, this did not help as much possibly because the model is already accustomed to predicting monolingual sentences and more resistant to switching.

\subsection{Accuracy on CS Switch-points}

The above experiments suggest that the \emph{CS-bias} loss is useful in improving CS ASR performance. However, it would be illustrative to examine the errors at code switching points and the influence of \emph{CS-bias} on these specific 
errors. \cite{karan_paper} introduces one such metric that helps us compute errors at the switching points. If there are $M$ words on both sides of the switch points across all reference transcriptions and $N$ of these switch point words are predicted correctly, then the code-switched WER (\textsc{CS-WER}) is defined as, \textsc{CS-WER} = $1 - \frac{N}{M}$.

Table~\ref{tab:cm_wer} lists the \textsc{CS-WER} values on both \devset and \testset datasets for a few important systems identified in the previous experiments. As intended, we see consistent improvements on the \textsc{CS-WER} metric with using the \emph{CS-bias} loss. Interestingly, the system with the best \textsc{CS-WER} score was one that used ``add-mixup" in a finetuning phase with a frozen encoder along with the \emph{CS-bias} loss. This could be because of the combination of \emph{CS-bias} explicitly encouraging switching and the frozen encoder preventing the model from encoding artefacts from TTS speech. 


\begin{table}[t!]
    \caption{Comparing \textsc{CS-WER} on different systems}
    \centering
    \begin{tabular}{l c c}
    \toprule
    \textbf{System}  & \textbf{DEV} & \textbf{TEST} \\
    \midrule
    Baseline & 82.4 & 83.3 \\
    \midrule
    no-mixup 100 (w/o \emph{CS-bias}) & 80.3 & 81.1 \\
    no-mixup 100 (with \emph{CS-bias}) & 79.7 & 81.2 \\
    add-mixup 100 (w/o \emph{CS-bias}) & 79.7 & 80.6 \\ 
    add-mixup 100 (with \emph{CS-bias}) & 79.3 & 79.8 \\
    \midrule
    add-mixup 100 (\textbf{FT}, w/o \emph{CS-bias}) & 79.1 & 80.4 \\
    add-mixup 100 (\textbf{FT}, with \emph{CS-bias}) & 78.7 & 79.8 \\
    add-mixup 100 (\textbf{FT}, with \textbf{FE}) & 75.5 & 78.3 \\
    add-mixup 100 (\textbf{FT}, \textbf{FE}, \emph{CS-bias}) & \textbf{73.1} & \textbf{76.1} \\
    \bottomrule
    \end{tabular}
    \label{tab:cm_wer}
\end{table}
    
\subsection{Finetuning on small amounts of real CS speech} 
\label{results:skyline}

In all the experiments so far, we assume no access to real CS speech and exclusively used synthetic CS speech (in addition to monolingual speech). However, a natural question that may arise is how our proposed techniques fare when the model can also make use of small amounts of real CS speech. To demonstrate this, we use speech from \csreal to finetune our baseline model. Table~\ref{tab:skyline} shows that even with access to just 12 hours of real CS speech the WERs  dramatically improve over the purely monolingual baseline system. Starting from the ``add-mixup 100" system that was trained with \emph{CS-bias} and further finetuning with \csreal further improves WERs on both \devset and \testset datasets. Table~\ref{tab:skyline} also shows the \textsc{CS-WER} values for all the models. Both our proposed techniques also help in achieving the best \textsc{CS-WER}.





\begin{table}[t!]
    \caption{WERs / CS-WERs after finetuning with real CS speech}    \centering
    \begin{tabular}{l c c}
    \toprule
    \textbf{System}     &  \textbf{DEV} & \textbf{TEST} \\
    \midrule
    Baseline (no CS speech) & 70.7/82.4 & 72.1 /83.3\\ 
    \midrule
    \textbf{FT} with \csreal (B1) & 58.6/59.8 & 59.9/60.2\\
    B1 + \emph{CS-bias} & 58.4/58.8 & 59.4/59.6\\
    \midrule
    A-100 (\textbf{FT} w/ \csreal) & 54.4/54.0 & 55.5/54.8 \\
    A-100 + \emph{CS-bias} (\textbf{FT} w/ \csreal) & \textbf{53.6}/\textbf{53.3} & \textbf{55.2}/\textbf{54.4}\\
    A-100 (\textbf{FT} w/ \csreal) + \emph{CS-Bias} & 54.8/54.7 & 55.7/55.5 \\
    \bottomrule
    \end{tabular}
    \label{tab:skyline}
\end{table}


\section{Conclusions}
In this work, we explore the usability of high quality text-to-speech (TTS) data as a resource for CS ASR in settings where natural code-switched speech is unavailable or is a scarce resource. We present two effective techniques to make use of TTS samples: mixup and a loss function that encourages code switching. We show performance improvements using both these techniques in two different experimental settings involving no real CS speech and small amounts of CS speech. This work opens up many interesting questions including whether there are other auxiliary loss functions that can help  code switching better and how the proposed techniques will scale to very large quantities of TTS samples. We leave these explorations for future work. 


\section{Acknowledgements}
We acknowledge Praneeth Lakmala and Arijit Mukherjee for their help in building multi-lingual Indic TTS models for the TTS data generation. We are also grateful to Niranjan Nayak, Rupesh Mehta, Sandeepkumar Satpal, Ankur Gupta and Satarupa Guha of Microsoft IDC for their time and support with hardware resources and corpora, and their valuable guidance.

\bibliographystyle{IEEEtran}

\bibliography{mybib}

\begin{thebibliography}{10}
\providecommand{\url}[1]{#1}
\csname url@samestyle\endcsname
\providecommand{\newblock}{\relax}
\providecommand{\bibinfo}[2]{#2}
\providecommand{\BIBentrySTDinterwordspacing}{\spaceskip=0pt\relax}
\providecommand{\BIBentryALTinterwordstretchfactor}{4}
\providecommand{\BIBentryALTinterwordspacing}{\spaceskip=\fontdimen2\font plus
\BIBentryALTinterwordstretchfactor\fontdimen3\font minus
  \fontdimen4\font\relax}
\providecommand{\BIBforeignlanguage}[2]{{%
\expandafter\ifx\csname l@#1\endcsname\relax
\typeout{** WARNING: IEEEtran.bst: No hyphenation pattern has been}%
\typeout{** loaded for the language `#1'. Using the pattern for}%
\typeout{** the default language instead.}%
\else
\language=\csname l@#1\endcsname
\fi
#2}}
\providecommand{\BIBdecl}{\relax}
\BIBdecl

\bibitem{zhang2017mixup}
H.~Zhang, M.~Cisse, Y.~N. Dauphin, and D.~Lopez-Paz, ``mixup: Beyond empirical
  risk minimization,'' 2017.

\bibitem{seame}
N.~T. Vu, D.-C. Lyu, J.~Weiner, D.~Telaar, T.~Schlippe, F.~Blaicher, E.-S.
  Chng, T.~Schultz, and H.~Li, ``A first speech recognition system for
  {Mandarin-English} code-switch conversational speech,'' in \emph{Proceedings
  of ICASSP}, 2012, pp. 4889--4892.

\bibitem{asymmetric-cs-asr}
Y.~Li, P.~Fung, P.~Xu, and Y.~Liu, ``Asymmetric acoustic modeling of mixed
  language speech,'' in \emph{Proceedings of ICASSP}, 2011, pp. 5004--5007.

\bibitem{univphoneme-mixed-asr}
D.~Imseng, H.~Bourlard, M.~M. Doss, and J.~Dines, ``Language dependent
  universal phoneme posterior estimation for mixed language speech
  recognition,'' in \emph{Proceedings of ICASSP}, 2011, pp. 5012--5015.

\bibitem{hkust-pointer-paper}
G.~I. Winata, A.~Madotto, C.~Wu, and P.~Fung, ``Learn to code-switch: Data
  augmentation using copy mechanism on language modeling,'' \emph{CoRR}, vol.
  abs/1810.10254, 2018.

\bibitem{FAME-paper}
E.~Yılmaz, H.~van~den Heuvel, and D.~Van~Leeuwen, ``Acoustic and textual data
  augmentation for improved {ASR} of code-switching speech,'' in
  \emph{Proceedings of Interspeech}, 2018, pp. 1933--1937.

\bibitem{adel2013recurrent}
H.~Adel, N.~T. Vu, F.~Kraus, T.~Schlippe, H.~Li, and T.~Schultz, ``Recurrent
  neural network language modeling for code switching conversational speech,''
  in \emph{Proceedings of ICASSP}, 2013, pp. 8411--8415.

\bibitem{csalt-dlm-speech-paper}
S.~Garg, T.~Parekh, and P.~Jyothi, ``Dual language models for code switched
  speech recognition,'' in \emph{Proceedings of Interspeech}, 2018, pp.
  2598--2602.

\bibitem{ect_paper}
A.~Pratapa, G.~Bhat, M.~Choudhury, S.~Sitaram, S.~Dandapat, and K.~Bali,
  ``Language modeling for code-mixing: The role of linguistic theory based
  synthetic data,'' in \emph{Proceedings of ACL}, 2018, pp. 1543--1553.

\bibitem{e2e-manglish-icassp19}
C.~{Shan}, C.~{Weng}, G.~{Wang}, D.~{Su}, M.~{Luo}, D.~{Yu}, and L.~{Xie},
  ``Investigating end-to-end speech recognition for {Mandarin-English}
  code-switching,'' in \emph{Proceedings of ICASSP}, 2019, pp. 6056--6060.

\bibitem{cm-asr-ctc-icassp19}
K.~{Li}, J.~{Li}, G.~{Ye}, R.~{Zhao}, and Y.~{Gong}, ``Towards code-switching
  {ASR} for end-to-end {CTC} models,'' in \emph{Proceedings of ICASSP}, 2019,
  pp. 6076--6080.

\bibitem{multilingual-and-cm-google-paper}
S.~Toshniwal, T.~N. Sainath, R.~J. Weiss, B.~Li, P.~J. Moreno, E.~Weinstein,
  and K.~Rao, ``Multilingual speech recognition with a single end-to-end
  model,'' in \emph{Proceedings of ICASSP}, 2018, pp. 4904--4908.

\bibitem{kim2016joint}
S.~Kim, T.~Hori, and S.~Watanabe, ``Joint {CTC}-attention based end-to-end
  speech recognition using multi-task learning,'' in \emph{Proceedings of
  ICASSP}, 2017, pp. 4835--4839.

\bibitem{srivastava2019endtoend}
B.~M.~L. Srivastava, B.~Abraham, S.~Sitaram, R.~Mehta, and P.~Jyothi,
  ``End-to-end {ASR} for code-switched {Hindi-English} speech,'' \emph{CoRR},
  vol. abs/1906.09426, 2019.

\bibitem{Yue2019EndtoEndCA}
X.~Yue, G.~Lee, E.~Yilmaz, F.~Deng, and H.~Li, ``End-to-end code-switching
  {ASR} for low-resourced language pairs,'' in \emph{Proceedings of ASRU},
  2019, pp. 972--979.

\bibitem{google-bytes-paper}
B.~Li, Y.~Zhang, T.~N. Sainath, Y.~Wu, and W.~Chan, ``Bytes are all you need:
  End-to-end multilingual speech recognition and synthesis with bytes,''
  \emph{CoRR}, vol. abs/1811.09021, 2018.

\bibitem{Ko2015AudioAF}
T.~Ko, V.~Peddinti, D.~Povey, and S.~Khudanpur, ``Audio augmentation for speech
  recognition,'' in \emph{Proceedings of Interspeech}, 2015, pp. 3586--3589.

\bibitem{SpecAugment}
D.~S. Park, W.~Chan, Y.~Zhang, C.-C. Chiu, B.~Zoph, E.~D. Cubuk, and Q.~V. Le,
  ``{SpecAugment: A Simple Data Augmentation Method for Automatic Speech
  Recognition},'' in \emph{Proceedings of Interspeech}, 2019, pp. 2613--2617.

\bibitem{Ragni2014DataAF}
A.~Ragni, K.~Knill, S.~P. Rath, and M.~J.~F. Gales, ``Data augmentation for low
  resource languages,'' in \emph{Proceedings of Interspeech}, 2014, pp.
  810--814.

\bibitem{merl-concat-paper}
H.~Seki, S.~Watanabe, T.~Hori, J.~L. Roux, and J.~R. Hershey, ``An end-to-end
  language-tracking speech recognizer for mixed-language speech,'' in
  \emph{Proceedings of ICASSP}, 2018, pp. 4919--4923.

\bibitem{karan_paper}
K.~Taneja, S.~Guha, P.~Jyothi, and B.~Abraham, ``Exploiting monolingual speech
  corpora for code-mixed speech recognition,'' in \emph{Proceedings of
  Interspeech}, 2019, pp. 2150--2154.

\bibitem{google-aug-tts}
A.~E. Rosenberg, Y.~Zhang, B.~Ramabhadran, Y.~Jia, P.~J. Moreno, Y.~Wu, and
  Z.~Wu, ``Speech recognition with augmented synthesized speech,'' in
  \emph{Proceedings of ASRU}, 2019, pp. 996--1002.

\bibitem{wang2020improving}
G.~Wang, A.~Rosenberg, Z.~Chen, Y.~Zhang, B.~Ramabhadran, Y.~Wu, and P.~Moreno,
  ``Improving speech recognition using consistent predictions on synthesized
  speech,'' in \emph{Proceedings of ICASSP}, 2020.

\bibitem{tts-aug-monolingual}
C.~{Du} and K.~{Yu}, ``Speaker augmentation for low resource speech
  recognition,'' in \emph{Proceedings of ICASSP}, 2020, pp. 7719--7723.

\bibitem{nara-chain-paper}
S.~Nakayama, A.~Tjandra, S.~Sakti, and S.~Nakamura, ``Speech chain for
  semi-supervised learning of {Japanese-English} code-switching {ASR} and
  {TTS},'' in \emph{SLT Workshop}, 2018, pp. 182--189.

\bibitem{jap-eng-tts-data}
S.~{Nakayama}, T.~{Kano}, Q.~T. {Do}, S.~{Sakti}, and S.~{Nakamura},
  ``Japanese-english code-switching speech data construction,'' in
  \emph{International Conference on Speech Database and Assessments}, 2018, pp.
  67--71.

\bibitem{kano2018structuredbased}
T.~Kano, S.~Sakti, and S.~Nakamura, ``Structured-based curriculum learning for
  end-to-end english-japanese speech translation,'' in \emph{Proceedings of
  Interspeech}, 2017, pp. 2630--2634.

\bibitem{DBLP:journals/corr/abs-1811-02050}
Y.~Jia, M.~Johnson, W.~Macherey, R.~J. Weiss, Y.~Cao, C.~Chiu, N.~Ari,
  S.~Laurenzo, and Y.~Wu, ``Leveraging weakly supervised data to improve
  end-to-end speech-to-text translation,'' in \emph{Proceedings of ICASSP},
  2019, pp. 7180--7184.

\bibitem{Medennikov2018AnIO}
I.~Medennikov, Y.~Y. Khokhlov, A.~Romanenko, D.~Popov, N.~A. Tomashenko,
  I.~Sorokin, and A.~Zatvornitsky, ``An investigation of mixup training
  strategies for acoustic models in {ASR},'' in \emph{Proceedings of
  Interspeech}, 2018, pp. 2903--2907.

\bibitem{saon2019sequence}
G.~Saon, Z.~T{\"u}ske, K.~Audhkhasi, and B.~Kingsbury, ``Sequence noise
  injected training for end-to-end speech recognition,'' in \emph{Proceedings
  of ICASSP}, 2019, pp. 6261--6265.

\bibitem{zhu2019mixup}
Y.~Zhu, T.~Ko, and B.~Mak, ``Mixup learning strategies for text-independent
  speaker verification,'' \emph{Proceedings of Interspeech}, pp. 4345--4349,
  2019.

\bibitem{liang2019levoice}
Y.~Liang, L.~Yang, X.~Wang, Y.~Li, C.~Jia, and J.~Wang, ``The levoice far-field
  speech recognition system for voices from a distance challenge 2019,''
  \emph{Proceedings of Interspeech}, pp. 2483--2487, 2019.

\bibitem{watanabe2018espnet}
S.~Watanabe, T.~Hori, S.~Karita, T.~Hayashi, J.~Nishitoba, Y.~Unno, N.~{Enrique
  Yalta Soplin}, J.~Heymann, M.~Wiesner, N.~Chen, A.~Renduchintala, and
  T.~Ochiai, ``{ESPnet}: End-to-end speech processing toolkit,'' in
  \emph{Proceedings of Interspeech}, 2018, pp. 2207--2211.

\bibitem{tacotron2}
J.~{Shen}, R.~{Pang}, R.~J. {Weiss}, M.~{Schuster}, N.~{Jaitly}, Z.~{Yang},
  Z.~{Chen}, Y.~{Zhang}, Y.~{Wang}, R.~{Skerrv-Ryan}, R.~A. {Saurous},
  Y.~{Agiomvrgiannakis}, and Y.~{Wu}, ``Natural {TTS} synthesis by conditioning
  {WaveNet} on {Mel} spectrogram predictions,'' in \emph{Proceedings of
  ICASSP}, 2018, pp. 4779--4783.

\bibitem{Vocoder}
A.~van~den Oord, Y.~Li, I.~Babuschkin, K.~Simonyan, O.~Vinyals, K.~Kavukcuoglu,
  G.~van~den Driessche, E.~Lockhart, L.~C. Cobo, F.~Stimberg, N.~Casagrande,
  D.~Grewe, S.~Noury, S.~Dieleman, E.~Elsen, N.~Kalchbrenner, H.~Zen,
  A.~Graves, H.~King, T.~Walters, D.~Belov, and D.~Hassabis, ``Parallel
  {WaveNet}: Fast high-fidelity speech synthesis,'' in \emph{Proceedings of
  ICML}, 2019.

\bibitem{sennrich2015neural}
R.~Sennrich, B.~Haddow, and A.~Birch, ``Neural machine translation of rare
  words with subword units,'' in \emph{Proceedings of ACL}, 2016, pp.
  1715--1725.

\bibitem{zeiler2012adadelta}
M.~D. Zeiler, ``Adadelta: an adaptive learning rate method,'' \emph{CoRR}, vol.
  abs/1212.5701, 2012.

\bibitem{amodei2015deep}
D.~Amodei, R.~Anubhai, E.~Battenberg, C.~Case, J.~Casper, B.~Catanzaro,
  J.~Chen, M.~Chrzanowski, A.~Coates, G.~Diamos, E.~Elsen, J.~Engel, L.~Fan,
  C.~Fougner, T.~Han, A.~Hannun, B.~Jun, P.~LeGresley, L.~Lin, S.~Narang,
  A.~Ng, S.~Ozair, R.~Prenger, J.~Raiman, S.~Satheesh, D.~Seetapun,
  S.~Sengupta, Y.~Wang, Z.~Wang, C.~Wang, B.~Xiao, D.~Yogatama, J.~Zhan, and
  Z.~Zhu, ``{Deep Speech 2}: End-to-end speech recognition in {English} and
  {Mandarin},'' in \emph{Proceedings of ICML}, 2016, p. 173–182.

\end{thebibliography}


\end{document}